\documentclass[10pt,twocolumn,letterpaper]{article}

\usepackage{cvpr}
\usepackage{times}
\usepackage{epsfig}

\usepackage{amsmath}
\usepackage{amssymb}

\usepackage{graphicx}
\usepackage{subcaption}
\usepackage[font=small,labelfont=bf]{caption}
\graphicspath{ {figures/} }

\newcommand{\norm}[1]{\left\lVert#1\right\rVert}

\usepackage{hyperref}

\cvprfinalcopy 

\begin{document}

\title{Using Deep Learning for Segmentation and Counting within Microscopy Data
	}

\author{Carlos Xavier Hern\'{a}ndez\\
Stanford University\\
Stanford, CA\\
{\tt\small cxh@stanford.edu}
\and
Mohammad M. Sultan\\
Stanford University\\
Stanford, CA\\
{\tt\small msultan@stanford.edu}
\and
Vijay S, Pande\thanks{Corresponding author: pande@stanford.edu}\\
Stanford University\\
Stanford, CA\\
{\tt\small pande@stanford.edu}
}

\maketitle

\begin{abstract}
Cell counting is a ubiquitous, yet tedious task that would greatly benefit from automation.
From basic biological questions to clinical trials, cell counts provide key quantitative
feedback that drive research. Unfortunately, cell counting is most commonly a manual task
and can be time-intensive. The task is made even more difficult due to overlapping cells,
existence of multiple focal planes, and poor imaging quality, among other factors.
Here, we describe a convolutional neural network approach, using a recently described
feature pyramid network combined with a VGG-style neural network, for segmenting and
subsequent counting of cells in a given microscopy image.
   
\end{abstract}

\section{Introduction}
Cell segmentation and counting can be a laborious task that can take up valuable time from research.
Typically, a scientist must manually estimate the number of cells in a local grid within an image~\cite{Abcam}.
This is repeated at various grid points across the plate to get a mean density which is then used for estimating the
total number of cells. These density-based techniques suffer from several drawbacks: first, they require a human
to manually count the number of cells, introducing the possibility of subjective errors;
second, they require a significant amount of time commitment, which could be better used for understanding, designing,
and performing a new series of experiments; finally, it is not completely obvious how error bars can be obtained from such
an analysis. Although more sophisticated tools do exist for this task, they can be costly, rely on closed-source software,
and do not address the issue of quantifying error.

Here, we aim to automate the tedious process of counting cells and estimating uncertainty using convolutional neural networks (CNNs).
Our methodology takes on a two-step approach:

\begin{itemize}\itemsep=2pt
\item Cell segmentation: Generate a mask capable of identifying cells in an image. 
\item Cell counting: Approximate a count and confidence interval from the mask generated in the prior step.
\end{itemize}

We demonstrate how our approach leads to reliable cell counting in our benchmark dataset.
Additionally, we discuss model interpretability and assess how our model learns to count.
Finally, we discuss how incorporating uncertainty into the counting problem can improve model
reliability and help identify possible failure cases.

\section{Previous Work}

\subsection{Segmentation}

Segmentation using CNNs is an important problem in computer vision and
significant progress has been made over the past few years~\cite{deepmask, sharpmask, fastrcnn, fasterrcnn, fpn2016, maskrcnn2017}.
Of particular note is the Mask-RCNN algorithm designed by Facebook AI Research (FAIR), which represents the current state-of-the-art
in the field \cite{maskrcnn2017}. In their manuscript, He, et al. describe the use of feature pyramid networks (FPN) to generate
smooth subject masks from a region of interest within an image \cite{maskrcnn2017, fpn2016}. The stated advantage of the FPN is
that it down-samples the input image several times to learn features at different scales, which is reported to yield improved
segmentation masks \cite{fpn2016}. As the accepted state-of-the art, we adopted this approach for the task of segmenting cells.

We note that cell segmentation using CNNs has been addressed at least once before in the literature. Van Valen, et al. developed DeepCell, which treats the segmentation task as classification problem on a pixel-by-pixel basis \cite{deepcell}. While successful at classifying different cell types in images, DeepCell produces fairly low-resolution segmentation masks and does not aim to solve the cell counting problem. Our approach is fundamentally different from DeepCell, as it leverages the FPN architecture for masking and cell counting.

\subsection{Counting}
Past work has attempted to count object densities from an image using a CNN \cite{hydra}.
In their work, O\~{n}oro-Rubio and L\'{o}pez-Sastre generate density maps from multiple image
patches, which they then sum over to calculate a final count. The authors report success in both counting
vehicles and pedestrians in separate benchmarks. While their approach is similar to ours in spirit, our model
predicts counts from high-contrast foreground masks, with no direct knowledge of the density.
This is in-part due to our dataset's limitations (see Section \ref{sec:data}), which only provides a binary foreground mask for select images.

\subsection{Uncertainty}

Understanding the uncertainty within a model is a critical part of generating better-informed predictions.
Unfortunately, deep neural networks are often used blindly and their limitations are ignored,
which can lead to unwanted consequences in practice \cite{bayes_uncer_2017}. In the case of counting cells,
we often rely on such assays for everything from basic research to clinical trials. The difference between
having a statistically significant result or not, in somes cases, has potential to save lives.
Quantifying error for these scientific problems can lead to improved statistical power and help resolve
experimental phenotypes more easily.

The recent work of Kendall and Gal details how to quantify uncertainty in CNN models and shows its importance in vision tasks \cite{bayes_uncer_2017}.
Even more interesting is how trivial it is to implement their derivation of a model's aleatoric uncertainty.
By simply modifying the $L^2$ loss function to include an uncertainty term, $\sigma$:

\begin{equation}\label{eq_uncer}
[\text{Aleatoric Loss}] = \frac{\norm{y - \hat{y}}_2}{2 \sigma} +  \log \sigma^2
\end{equation}

where $\hat{y}$ is the model prediction and $y$ is the ground truth.

The aleatoric uncertainty of a model captures its uncertainty with respect to information which the training data cannot explain.
In the idealize case of a complete training dataset and overfit model, the aleatoric uncertainty of a given model input would be zero.
Therefore, learning the aleatoric uncertainty can yield some insights into what the model has not seen before at test-time
and, therefore, can inform the experimentalist about what training examples may need to be added in future datasets.

\section{Technical Approach}
\subsection{Cell Segmentation using a Feature Pyramid Network}

\begin{figure}[!h]
\centering
\includegraphics[width=0.5\textwidth]{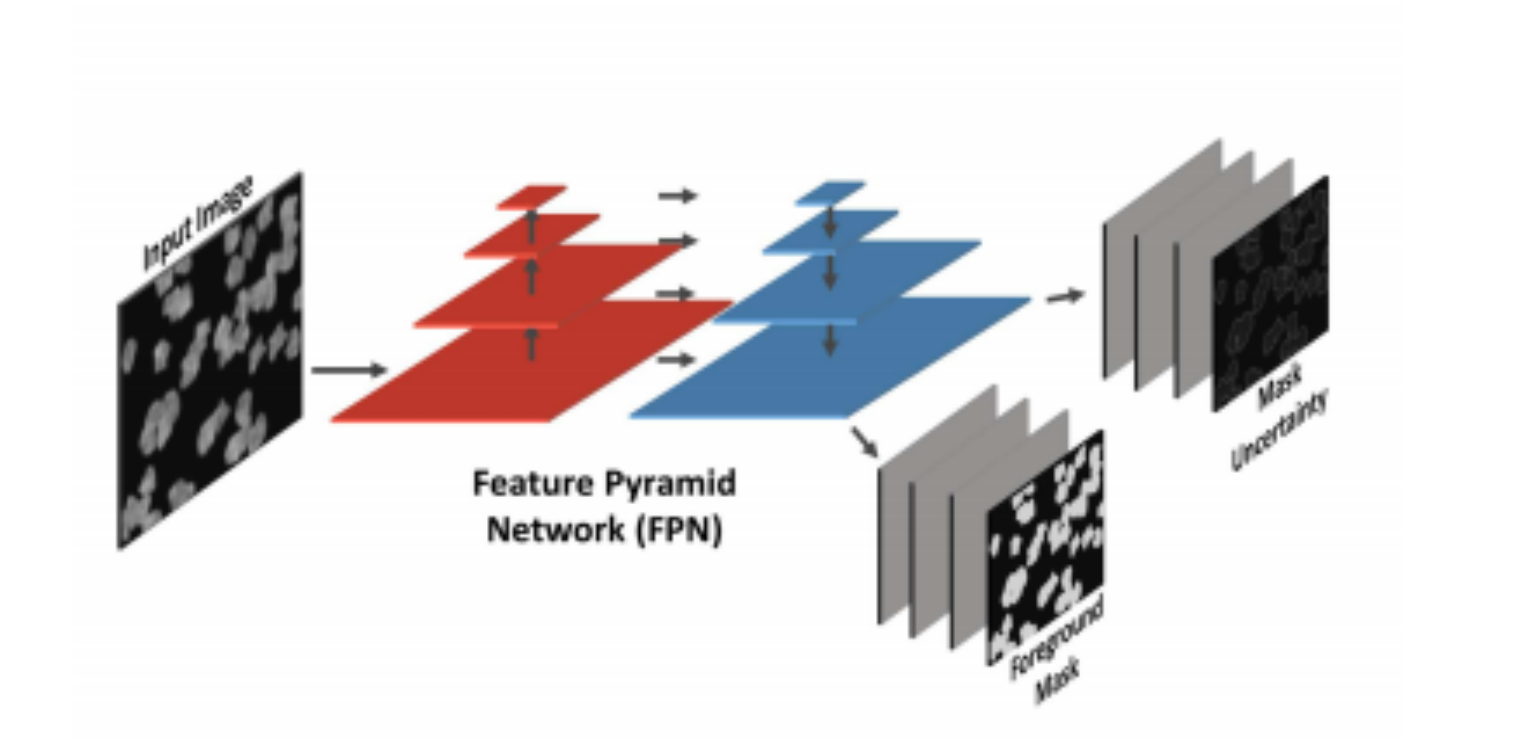}
\caption{A schematic of our Feature Pyramid Network for generating a foreground mask. An input image is first passed through the down-sampling network (red) which successively convolves and scales  down the image by a factor of $2$ within each layer. Next, the down-sampled image is passed through an up-sampling network (blue), which also convolves the up-sampled image with its corresponding scaled down-sampled image in the red network, as denoted by the arrows. Finally, the output of the last layer in up-sampling network
is passed through two separate networks: one that predicts the mean foreground masks and another that predicts its associated aleatoric uncertainty.}
\label{fig1}
\end{figure}

We begin by using an FPN to segment foreground cells from their background.
The FPN is a feature extraction network designed to build feature maps at multiple spatial scales \cite{fpn2016}.
It is a computationally efficient algorithm that cross-links a convolved down-sampled image with its corresponding up-sampled image in the network, as shown in Fig. \ref{fig1}. The cross link uses a single 3x3 convolution layer with 128 filters and padding to preserve the down-sampled images' dimensions. 
These cross-links enable the network to infer not only the relevant features at different scales but their spatial correlations as well.
Finally each up-sampled layer, goes through three 3x3 convolutional layers: the first two have 256 filters and the final one only has one to produce a black-and-white foreground mask.
All convolutional layers were proceeded by batch-normalization apart from the final output layer. The result is an arbitrary number of foreground masks at different scales from the original image.
For our network, we down-sample each layer in the pyramid by a factor of $2$ and have a pyramid depth of 4 resulting in four output masks, the largest of which is half the size of the original image and is the mask used later by the counting network.
A separate, yet similar, set of convolutional layers is used to generate the mask uncertainties, expressed as log-variances, as shown in Fig. \ref{fig1}.

Each mask has its associated loss function, which contains two terms: 1) the aleatoric loss from the true foreground and 2) the total-variational (TV) loss. The aleatoric loss, as described before, behaves as an $L^2$ loss funtion with the addition of the mask uncertainty. The TV loss helps to smooth the foreground mask and remove any unwanted artifacts in a semi-unsupervised fashion~\cite{tvloss}. The sum of these losses, for each scaled masked, is then used as the final loss for the FPN model.

\subsection{Cell Counting using a VGG-like Network}

\begin{figure}[!h]
\centering
\includegraphics[width=0.5\textwidth]{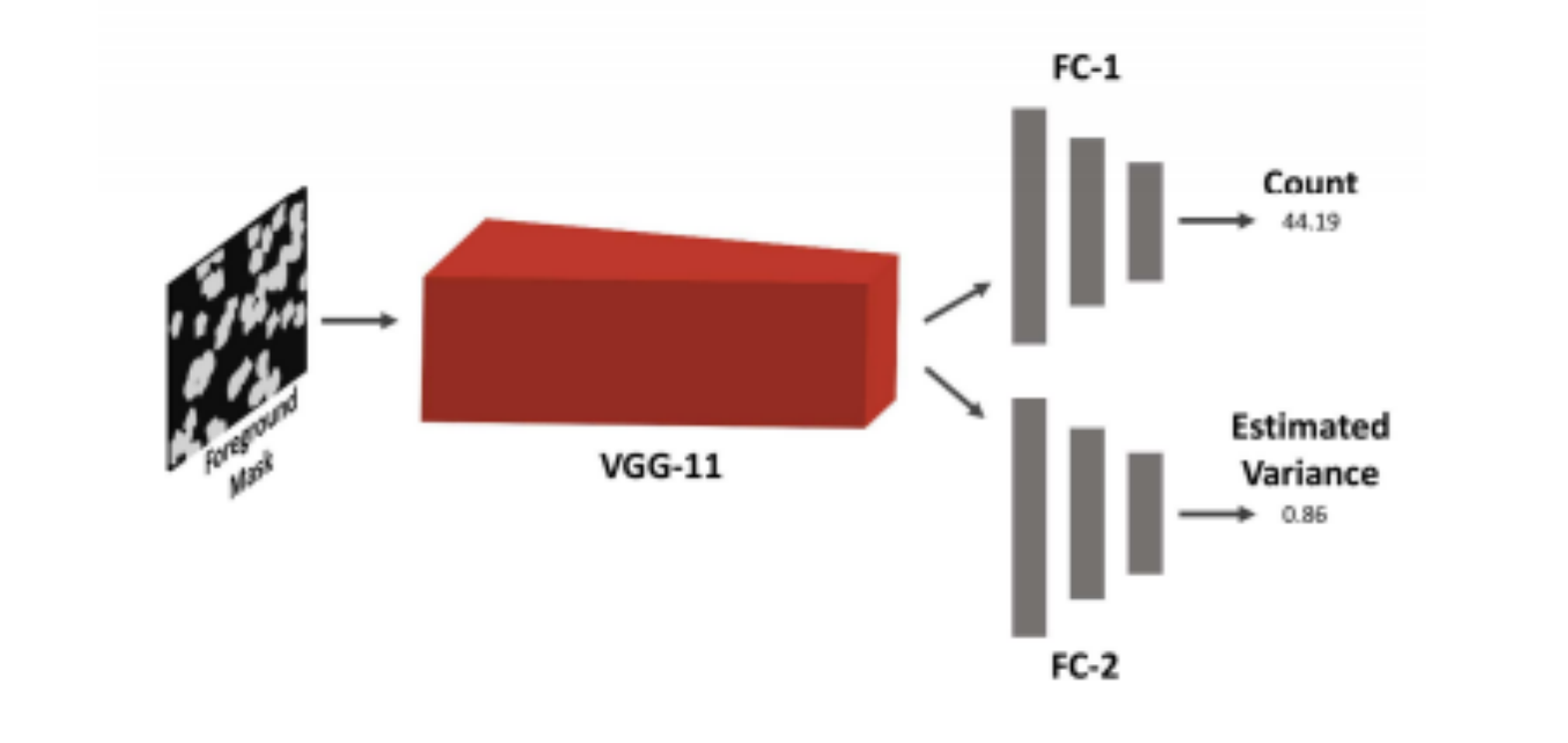}
\caption{A schematic of our VGG-11-based network for counting cells from a foreground mask generated using the network from Figure \ref{fig1}. The predicted foreground mask is passed through the VGG-11 architecture, after which outputs into a pair of fully-connected layers that predict the expected cell count and its associated aleatoric uncertainty.}
\label{fig2}
\end{figure}

For the counting network, we refer to the recent work of the Visual Geometry Group (VGG) in creating state-of-the-art, multi-purpose CNNs~\cite{vgg}. VGG networks are deep convolutional neural networks that, in the past, have won the ImageNet Challenge \cite{vgg}.
We chose to use the VGG-11 network which consists of 11 layers of two-dimensional convolutions with a filter size of 3x3 pixels.
The number of filters vary from 64 to 512 as the network progresses. Furthermore, each convolutional layer was also followed by a batch-normalization layer, which has been shown to improve training stability by preventing numerical instabilities in gradient calculations \cite{batchnorm}. This was followed by a leaky rectified linear unit (ReLU) activation and a max-pooling layers.

The VGG-11 network was used as a feature extractor for the counting network. To generate the count prediction, we used three fully connected layers which were again separated by a batch-normalization layer and leaky ReLu. These fully connected layers had dimension 1024, 512, and 1, respectively, to end with a single float. The final layer included a ReLU layer to prevent the network from outputting negative counts. A similar fully connected network, without the final ReLU layer, was used to predict the log-variance.
A general schematic of the final network is shown in Fig. \ref{fig2}.

\section{Dataset}
\label{sec:data}
\begin{figure}[!t]
\centering
\includegraphics[width=0.45\textwidth]{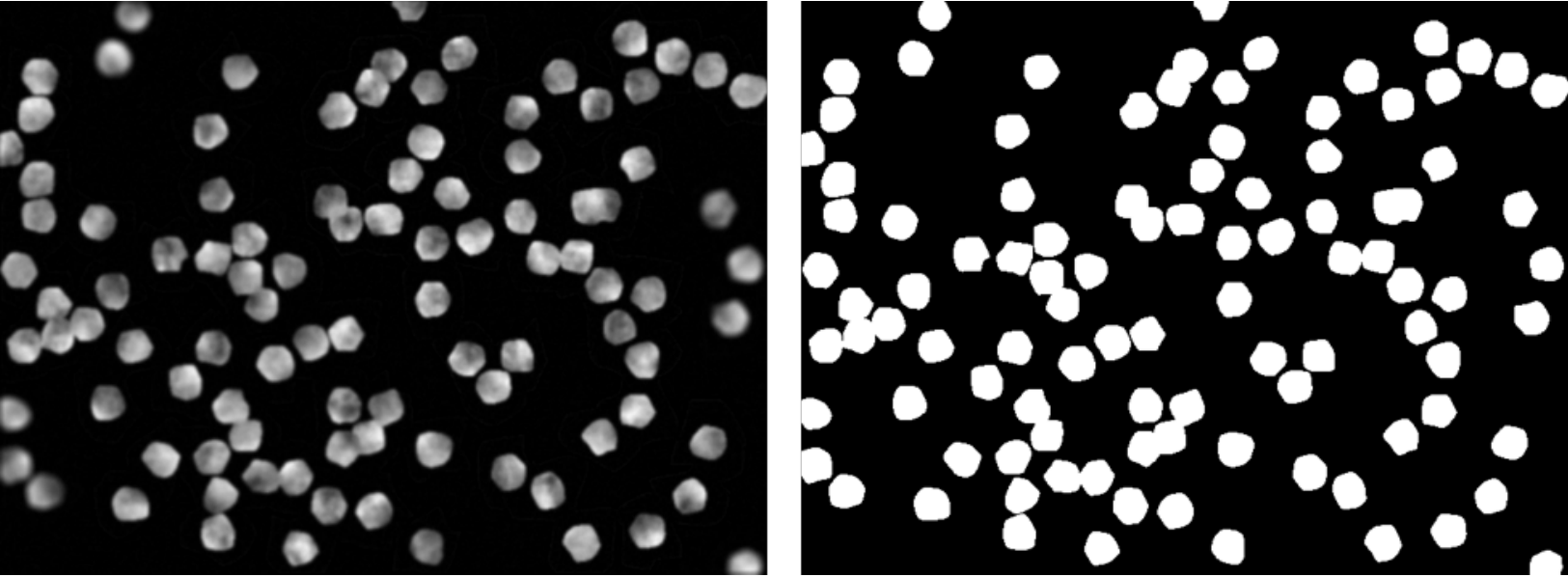}
\caption{Sample in-focus image from the Broad dataset. The raw image is shown on the left while the masked image is shown on the right.}
\label{fig_broad}
\end{figure}

We used the \href{https://data.broadinstitute.org/bbbc/BBBC005/}{BBBC005 dataset} from the Broad Institute's Bioimage Benchmark Collection \cite{bbbc}. This dataset is a collection of 9,600 simulated microscopy images of cell body-stained cells. An example is shown in Fig. \ref{fig_broad}.

These images were simulated using the SIMCEP simulation platform for a given cell count with a clustering probability of 25\% and cell areas matched to the average cell areas of human U2OS cells \cite{simcep1, simcep2}.
Focal blur was simulated by applying variable Gaussian filters to the images. Each image is 696 x 520 pixels encoded in the 8-bit Tagged Image File Format.
However, for the purposes of our experiment, images were eventually converted to JPEG format and scaled down to 256 x 192 pixels.

Of the 9,600 images, 600 images have a corresponding foreground mask. All 9600 images have associated cell counts with an upper limit of 100. The FPN was  trained on the 600 images while the Counting network was trained on the full dataset. We use about 100 of those for fast prototyping. We used a standard 80-20 train/test split for the final models.

\section{Results}

\begin{figure}[!h]
\centering
   \begin{subfigure}[b]{0.48\textwidth}
   \includegraphics[width=1.\textwidth]{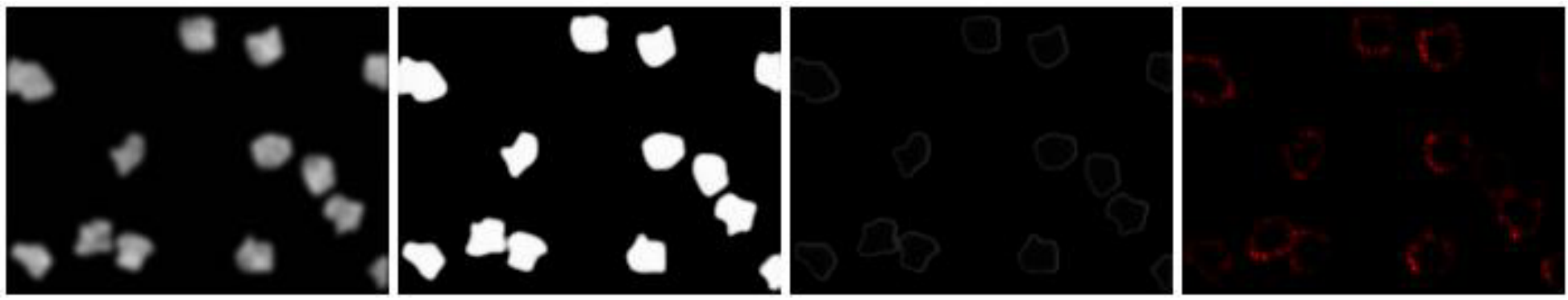}
   \caption{}
   \label{fig_sal}
   \end{subfigure}

   \begin{subfigure}[b]{0.48\textwidth}
   \includegraphics[width=1.\textwidth]{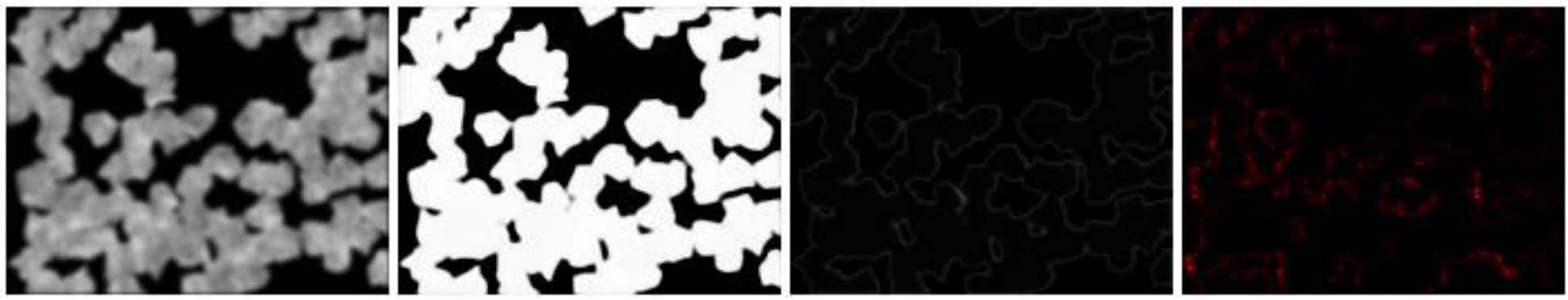}
   \caption{}
   \label{fig_sal2}
   \end{subfigure}
\caption{Examples where our model successfully predicted the number of cells in an image within its expected 95\% confidence interval. The left-most image is the input to our model.
To its right is the predicted foreground mask for the input and its associated uncertainty, respectively.
The right-most image depicts the saliency map during counting. ({\bf a}) A test example from our model with 14 cells.
Our model predicts that this image has 14.00 $\pm$ 1.82 number of cells with 95\% confidence.
({\bf b}) Another test example from our model with 96 cells. Our model predicts that this image has 96.17 $\pm$ 3.62 cells.}
\label{fig_good}
\end{figure}

\subsection{Segmentation}
We pre-trained the FPN on 480 images with known foregrounds, out of a total 600 images with this information.
Training was done using the sum of $L^2$ loss functions with aleatoric uncertainties and TV losses for each output mask with its associated uncertainty, as described previously, and optimized using the ADAM optimizer with a learning rate of $1E-3$
and batch size of 2.

We found that the average mean-squared error (MSE) on 100 validation images converged to a value less than $0.1$ after about 50 epochs.
Visual inspection of the masks over each epoch also corroborated this convergence, as masks looked nearly identical to the ground truth.
Convergence of the uncertainty masks also occurred over the 50 epochs, evolving from uniform uncertainty over the image to outlines of the cell clusters.

\subsection{Counting}

After training the FPN, we trained the VGG-11-based network on 7,680 FPN-generated masks out of 9,600 total images in our dataset.
Training was done using an $L^2$ loss function with aleatoric uncertainty for cell counts, optimizing with the ADAM optimizer with a learning rate of $1E-4$ and batch size of 5.

We found that the average validation MSE of the counts from 1,420 seperately generated masks converged after after 50 epochs
to a final value of less than 11.2.
After training, we found that our best model is able to achieve an $R^2$ value of $.987$, with an average $L^1$ error of 2.4 cells, on our test set. Furthermore, when considering the uncertainty predictions of our model, we find that over 80\% of ground truth counts fall within the model's predicted 95\% confidence interval on our 500 image test set. The maximum $L^1$ error never exceeded six cells during testing, which seems fairly accurate for the task.

Fig. \ref{fig_sal} and Fig. \ref{fig_sal2} demonstrate two random examples from our test set for which our model was successful.

\subsection{Saliency Mapping}
An unfortunate result of the complexity of modern deep learning models is that they can be difficult to interpret reliably. Good model interpretation can increase our understanding of the underlying problem, highlighting potentially insightful non-trivial patterns within datasets. These insights might be useful for designing future experiments or perhaps even improving the model itself.  

In this vein, we decided to probe our model further using saliency mapping \cite{saliency}. These maps are a standard technique in CNN literature to probe the internal states of the neural network. At a simplistic level, they are designed to highlight pixels in the data that maximally influence the predicted score. Saliency maps tend to highlight important features within the data which can then be used to understand what it is that the model is maximally looking at. 

We applied the saliency mapping technique to the counting network loss. The results are shown in Fig. \ref{fig_good} and Fig. \ref{fig_bad}. Our analysis shows that the network identifies outlines of cells within individual images. More importantly, the saliency map results seem agnostic to the number, size, and, orientation of cells within the images, emphasizing the general applicability of the model.

\subsection{Failure Cases}

In reviewing the saliency maps where the model fails to predict the correct number of cells within an expected confidence interval, we find three cases where our model seems to systematically fails:

\begin{itemize}\itemsep=1pt
\item High cell overlap
\item Irregular cell shapes 
\item Bad focal planes
\end{itemize}

\begin{figure}[!h]
\centering
   \begin{subfigure}[b]{0.48\textwidth}
   \includegraphics[width=1.\textwidth]{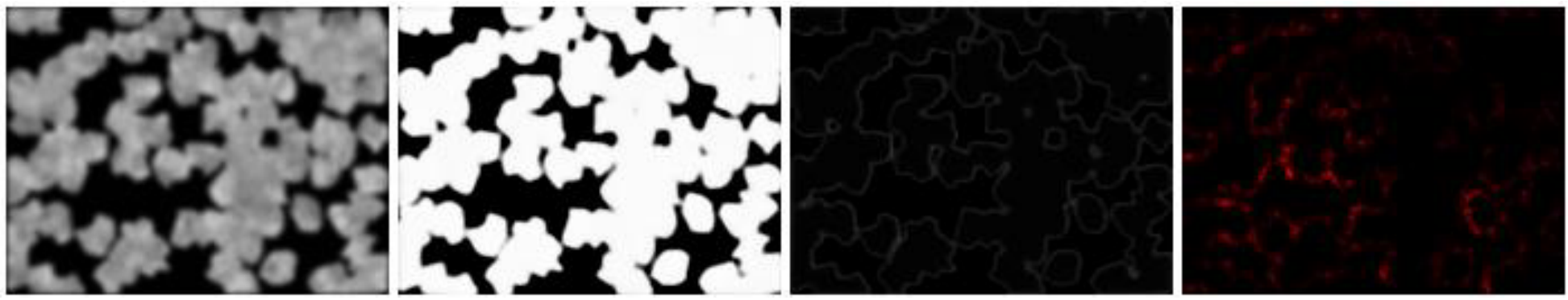}
   \caption{}
   \label{fig_bad1}
   \end{subfigure}

   \begin{subfigure}[b]{0.48\textwidth}
   \includegraphics[width=1.\textwidth]{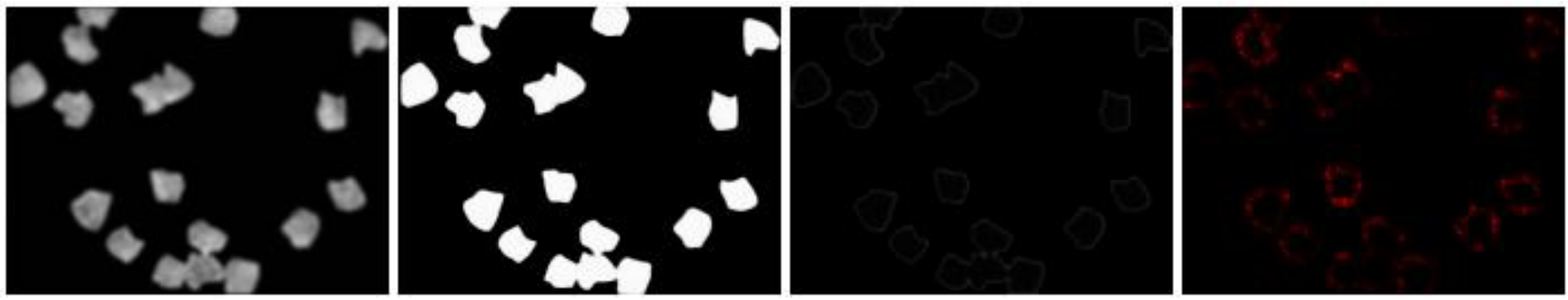}
   \caption{}
   \label{fig_bad2}
   \end{subfigure}
   
   \begin{subfigure}[b]{0.48\textwidth}
   \includegraphics[width=1.\textwidth]{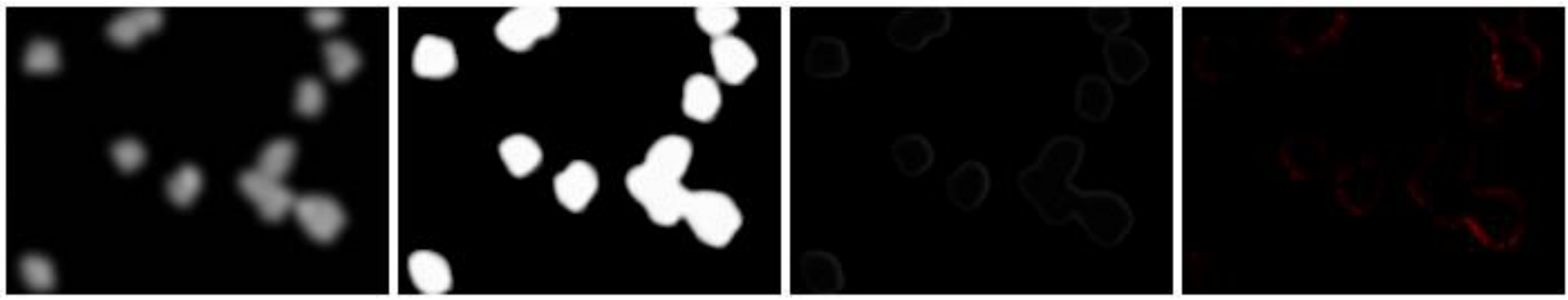}
   \caption{}
   \label{fig_bad3}
   \end{subfigure}
\caption{A few examples where our model successfully predicted the number of cells in an image within its expected 95\% confidence interval, due to systematic biases. The left-most image is the input to our model.
To its right is the predicted foreground mask for the input and its associated uncertainty, respectively.
The right-most image depicts the saliency map during counting. ({\bf a}) A test image with 100 cells with lots of overlapping. Our model predicts that this image has 93.50 $\pm$ 3.51 cells.
({\bf b}) A test image with 18 cells, a few of which are oddly shaped. Our model predicts that this image has 22.83 $\pm$ 2.19 cells.
({\bf c}) A test image with 14 cells taken at a poor focal plane. Our model predicts that this image has 11.85 $\pm$ 1.77 cells.}
\label{fig_bad}
\end{figure}

Examples of each case are demonstrated in Fig. \ref{fig_bad1}, Fig. \ref{fig_bad2}, and Fig. \ref{fig_bad3}, respectively.

In Fig. \ref{fig_bad1}, the input image has a number of regions with high cell density, which make it difficult even for a human to count. From the saliency map, we find that, although it recognizes many of the smaller patches of cells, the model ignores a large mass of cells in the upper right-hand corner of the image, causing it to undercount. We assess that this is possibly due to the model being unable to find a satisfiable edge to count in that region.

In Fig. \ref{fig_bad2}, the input image has a patch of cells near the center with an irregular shape. Although the image only has 18 cells, our model predicts that it should have at least 21. Upon further inspection of the saliency map, we assess that the model is counting multiple edges in the irregularly shaped patch, as irregularities in these cells are highlighted along with the exterior edges that we normally find highlighted.

In Fig. \ref{fig_bad3}, the input image is simulated in a bad focal plane, yielding significant blurring. In this case, the FPN does a poor job at creating a reliable foreground mask, which leads to undercounting. We find in the saliency map that a number of cell edges are ignored by our model.

\section{Discussion}
We show that it is possible to design and train a CNN architecture to count cells in microscopy images
and achieve relatively good accuracy. While a number of failure cases do arise, we believe that
better training data might help to overcome these shortcomings. Specifically, the BBBC005 lacked foreground masks
for out-of-focus images, which could have helped greatly in improving counting performance. However, some cases might
continue to persist, such as the issue of overlapping cells. At the moment, is unclear how to design a model to overcome
such a difficulty, as overlapping cells can even confound human experts, but one solution might be
to include the original image as additional input to the counting network. Another solution might be taking randomly
cropped image patches and robustly estimating the count from the average density over the many patches,
similar to O\~{n}oro-Rubio and L\'{o}pez-Sastre's approach \cite{hydra}.

The failure cases we discovered also raise the question as to whether human counters mightperform any better than our
algorithm and highlight the need for datasets that include expert predictions in this space. In lacking such a dataset,
we are currently unable to assess whether or not our alogrithm is on par with human performance.

Finally, we demonstrate a few good use-cases for aleatoric losses in estimating uncertainty
in cell counting. For one, it helps us define failure cases as instances when the ground truth lies outside of some acceptable tolerance.
In carefully defining failures, we were able to identify specific systematic cases of failure, which we can improve upon in future work.
As the eventual goal is to create a useful scientific tool, generating error bounds is
essential, as it improves the statistical power of our method and yields the ability to form
better informed hypotheses.

\section{Future Work}
Several extensions to the work presented are possible. On the methodology side, we have not yet worked extensively in optimizing the depth or architecture of the counting network.
It is entirely possibly that other variants of the VGG network or even more modern networks, such as Residual Networks, might yield better performance \cite{resnet}.
Furthermore, fully applying transfer learning to our FPN towards the counting task, where we would fine-tune the model during the
counting, might improve our final results.

We are also interested in training on more real-world datasets and have our model compete with human experts.
We believe that this will yield more insight into the counting problem and better diagnose the limitations of our model compared to human performance.
Not only this, but we strongly believe that any additional datasets can only improve our model's performance, as certain aspects of BBBC005 were not ideal.

Another way we might be able to leverage more real-world datasets is by using our network to tackle visual reasoning in microscopy images,
similar to the work of Johnson, et al. \cite{visreas}. One might imagine a researcher asking a visual reasoning algorithm to mask and count
specific kinds of cells in practice. This would greatly accelerate the pace of biological research and reduce the need for humans to interpret
microscopy data altogether.

On the applications side, we hope to build the current models into an easy-to-use smart phone or web application, allowing researchers to freely use our model in their research.
We also hope to integrate the ability for the research to submit their own training examples to help improve our model.
Open-source microscopy platforms like Foldscope might offer a great source of training images and community involvement to get started \cite{foldscope}.
Ultimately, we believe cell counting should evolve from a tedious manual task to a fully automated process simply using relatively cheap digital imaging and computing resources. 

\section*{Availability}
Source code for this work is available under the open-source
MIT license and is accessible on GitHub: \href{https://github.com/cxhernandez/cellcount}{https://github.com/cxhernandez/cellcount}.

\section*{Acknowledgements}
We would like to thank Bharath Ramsundar and Timnit Gebru for helpful conversations and feedback.
We acknowledge funding from NIH grants U19 AI109662 and 2R01GM062868 for their support of this work.
CXH acknowledges support from NSF GRFP (DGE-114747). 
M.M.S would like to acknowledge support from the National Science Foundation grant NSF-MCB-0954714.
This work used the XStream computational resource, supported by the National Science Foundation Major Research Instrumentation program (ACI-1429830), as well as the Sherlock cluster, maintained by the Stanford Research Computing Center.

\section*{Disclosures}
VSP is a consultant and SAB member of Schrodinger, LLC and Globavir, sits on the
Board of Directors of Apeel Inc, Freenome Inc, Omada Health, Patient Ping,
Rigetti Computing, and is a General Partner at Andreessen Horowitz.

{\small
\bibliographystyle{ieee}
\bibliography{egbib}

\begin{thebibliography}{10}\itemsep=-1pt

\bibitem{Abcam}
Counting cells using a hemocytometer.
\newblock
  \url{http://www.abcam.com/protocols/counting-cells-using-a-haemocytometer}.
\newblock Accessed: 2017-06-09.

\bibitem{foldscope}
J.~S. Cybulski, J.~Clements, and M.~Prakash.
\newblock Foldscope: Origami-based paper microscope.
\newblock {\em {PLoS} {ONE}}, 9(6):e98781, jun 2014.

\bibitem{fastrcnn}
R.~B. Girshick.
\newblock Fast {R-CNN}.
\newblock {\em CoRR}, abs/1504.08083, 2015.

\bibitem{maskrcnn2017}
K.~He, G.~Gkioxari, P.~Doll{\'{a}}r, and R.~B. Girshick.
\newblock Mask {R-CNN}.
\newblock {\em CoRR}, abs/1703.06870, 2017.

\bibitem{resnet}
K.~He, X.~Zhang, S.~Ren, and J.~Sun.
\newblock Deep residual learning for image recognition.
\newblock {\em CoRR}, abs/1512.03385, 2015.

\bibitem{batchnorm}
S.~Ioffe and C.~Szegedy.
\newblock Batch normalization: Accelerating deep network training by reducing
  internal covariate shift.
\newblock {\em CoRR}, abs/1502.03167, 2015.

\bibitem{tvloss}
M.~Javanmardi, M.~Sajjadi, T.~Liu, and T.~Tasdizen.
\newblock Unsupervised total variation loss for semi-supervised deep learning
  of semantic segmentation.
\newblock {\em CoRR}, abs/1605.01368, 2016.

\bibitem{visreas}
J.~Johnson, B.~Hariharan, L.~van~der Maaten, J.~Hoffman, F.~Li, C.~L. Zitnick,
  and R.~B. Girshick.
\newblock Inferring and executing programs for visual reasoning.
\newblock {\em CoRR}, abs/1705.03633, 2017.

\bibitem{bayes_uncer_2017}
A.~Kendall and Y.~Gal.
\newblock What uncertainties do we need in bayesian deep learning for computer
  vision?
\newblock {\em CoRR}, abs/1703.04977, 2017.

\bibitem{simcep1}
A.~Lehmussola, P.~Ruusuvuori, J.~Selinummi, H.~Huttunen, and O.~Yli-Harja.
\newblock Computational framework for simulating fluorescence microscope images
  with cell populations.
\newblock {\em {IEEE} Transactions on Medical Imaging}, 26(7):1010--1016, jul
  2007.

\bibitem{simcep2}
A.~Lehmussola, P.~Ruusuvuori, J.~Selinummi, T.~Rajala, and O.~Yli-Harja.
\newblock Synthetic images of high-throughput microscopy for validation of
  image analysis methods.
\newblock {\em Proceedings of the {IEEE}}, 96(8):1348--1360, aug 2008.

\bibitem{fpn2016}
T.~Lin, P.~Doll{\'{a}}r, R.~B. Girshick, K.~He, B.~Hariharan, and S.~J.
  Belongie.
\newblock Feature pyramid networks for object detection.
\newblock {\em CoRR}, abs/1612.03144, 2016.

\bibitem{bbbc}
V.~Ljosa, K.~L. Sokolnicki, and A.~E. Carpenter.
\newblock Annotated high-throughput microscopy image sets for validation.
\newblock {\em Nature Methods}, 9(7):637--637, jun 2012.

\bibitem{hydra}
D.~O\~noro Rubio and R.~J. L\'opez-Sastre.
\newblock Towards perspective-free object counting with deep learning.
\newblock In {\em ECCV}, 2016.

\bibitem{deepmask}
P.~O. Pinheiro, R.~Collobert, and P.~Dollár.
\newblock Learning to segment object candidates.
\newblock In {\em NIPS}, 2015.

\bibitem{sharpmask}
P.~O. Pinheiro, T.-Y. Lin, R.~Collobert, and P.~Dollár.
\newblock Learning to refine object segments.
\newblock In {\em ECCV}, 2016.

\bibitem{fasterrcnn}
S.~Ren, K.~He, R.~B. Girshick, and J.~Sun.
\newblock Faster {R-CNN:} towards real-time object detection with region
  proposal networks.
\newblock {\em CoRR}, abs/1506.01497, 2015.

\bibitem{saliency}
K.~Simonyan, A.~Vedaldi, and A.~Zisserman.
\newblock Deep inside convolutional networks: Visualising image classification
  models and saliency maps.
\newblock {\em CoRR}, abs/1312.6034, 2013.

\bibitem{vgg}
K.~Simonyan and A.~Zisserman.
\newblock Very deep convolutional networks for large-scale image recognition.
\newblock {\em CoRR}, abs/1409.1556, 2014.

\bibitem{deepcell}
D.~A.~V. Valen, T.~Kudo, K.~M. Lane, D.~N. Macklin, N.~T. Quach, M.~M.
  DeFelice, I.~Maayan, Y.~Tanouchi, E.~A. Ashley, and M.~W. Covert.
\newblock Deep learning automates the quantitative analysis of individual cells
  in live-cell imaging experiments.
\newblock {\em {PLOS} Computational Biology}, 12(11):e1005177, nov 2016.

\end{thebibliography}
}

\end{document}